\documentclass{article}
\usepackage{ijcai26}

\usepackage{times}
\usepackage{soul}
\usepackage{url}
\usepackage[hidelinks]{hyperref}
\usepackage[utf8]{inputenc}
\usepackage[small]{caption}
\usepackage{graphicx}
\usepackage{amsmath}
\usepackage{amsthm}
\usepackage{booktabs}
\usepackage{algorithm}
\usepackage{algorithmic}
\usepackage{amsfonts}
\usepackage{tabularx}
\usepackage{multirow}
\usepackage{pifont}
\newcommand{\cmark}{\ding{51}}%
\newcommand{\xmark}{\ding{55}}%
\usepackage[table]{xcolor}
\usepackage[switch]{lineno}

\title{X-Mark: Saliency-Guided Robust Dataset Ownership Verification \\for Medical Imaging}

\author{
Pranav Kulkarni\and
Junfeng Guo\And
Heng Huang\\
\affiliations
University of Maryland Institute for Health Computing, North Bethesda, MD\\
\emails
\{pranavk, gjf2023, heng\}@umd.edu,
}

\begin{document}

\maketitle

\begin{abstract}
High-quality medical imaging datasets are essential for training deep learning models, but their unauthorized use raises serious copyright and ethical concerns. Medical imaging presents a unique challenge for existing dataset ownership verification methods designed for natural images, as static watermark patterns generated in fixed-scale images scale poorly dynamic and high-resolution scans with limited visual diversity and subtle anatomical structures, while preserving diagnostic quality. In this paper, we propose X-Mark, a sample-specific clean-label watermarking method for chest x-ray copyright protection. Specifically, X-Mark uses a conditional U-Net to generate unique perturbations within salient regions of each sample. We design a multi-component training objective to ensure watermark efficacy, robustness against dynamic scaling processes while preserving diagnostic quality and visual-distinguishability. We incorporate Laplacian regularization into our training objective to penalize high-frequency perturbations and achieve watermark scale-invariance. Ownership verification is performed in a black-box setting to detect characteristic behaviors in suspicious models. Extensive experiments on CheXpert verify the effectiveness of X-Mark, achieving WSR of $100\%$ and reducing probability of false positives in Ind-M scenario by $12\%$, while demonstrating resistance to potential adaptive attacks.

\end{abstract}

\section{Introduction}

The release of large-scale, high-quality medical imaging datasets has advanced the utility of deep learning (DL) models in clinical settings, demonstrating its potential across diagnosis, prognosis, and treatment planning \cite{rajpurkar2023current}. Curating such datasets is a substantial investment, both time-consuming and expensive, especially when expert annotations from radiologists are required \cite{diaz2024monai,ma2024segment}. The value of these high-quality datasets is further amplified as radiology AI tools are rapidly commercialized and cleared by the FDA \cite{ebrahimian2022fda}. This raises concerns regarding the unauthorized commercial use of public datasets released solely for research purposes. 

Consider the MIMIC dataset, a large publicly accessible collection of chest x-rays (CXRs), radiology reports, and electronic health record \cite{johnson2019mimic,johnson2023mimic}. It represents a high-value resource for research, requiring a data use agreement and credentialing process involving HIPAA training. Despite these safeguards, the dataset remains susceptible to unauthorized commercial use if distributed outside of approved channels. Such misuse is not only a serious violation of the dataset owners' intellectual property, but also presents significant ethical concerns regarding patient privacy.

While several groups have explored dataset copyright protection for DL models, they are primarily designed for natural images, and methods intended for medical imaging remain relatively underexplored \cite{abadi2016deep,hua2023unambiguous}. Dataset ownership verification (DOV) is the most common and effective strategy, relying on the embedding of watermarks into a small subset of samples within the released dataset \cite{li2022untargeted,li2023blackbox}. Any model trained on this watermarked dataset learns the association between the watermark and specific labels, allowing the dataset owner to detect unauthorized use even in a black-box setting.

In this work, we focus on CXRs which present two challenges for existing DOV methods. First, previous methods rely on fixed-size static backdoor patterns, which scale poorly to the dynamic and high-resolution scans typically encountered with CXRs (e.g., often exceeding $2000 \times 2500$). Since DL models are trained on downsampled versions of these images, watermarks must remain effective even after resizing. Second, CXRs exhibit limited visual diversity and subtle anatomical structures in grayscale, requiring watermarks to be embedded within a single-channel while preserving diagnostic quality. This constraint makes watermark effectiveness and imperceptibility especially challenging: If the perturbation is too strong, watermarks may become easy to detect upon manual inspection. High-frequency perturbations that are imperceptible may not survive downsampling, while perturbations that blend with the image characteristics may inadvertently trigger benign models. Together, these challenges limit the effectiveness of DOV methods in medical imaging.

To address these limitations, we propose X-Mark, a sample-specific clean-label backdoor watermarking method for CXR copyright protection. Our method uses a U-Net with an EigenCAM-based saliency conditioning module that generates unique perturbations within salient regions of each sample. We design a multi-component training objective to ensure watermark efficacy, robustness against dynamic scaling processes, while preserving diagnostic quality and visually-distinguishability. We incorporate Laplacian regularization \cite{denton2015deep} into our training objective to penalize high-frequency perturbations and achieve watermark scale-invariance. Extensive experiments on CheXpert \cite{irvin2019chexpert} verify the effectiveness of X-Mark, achieving WSR of $100\%$ and reducing probability of false positives in Ind-M scenario by $12\%$, while demonstrating resistance to potential adaptive attacks. Our main contributions are three-fold:
\begin{itemize}
    \item We propose a sample-specific clean-label backdoor watermarking method for medical image copyright protection. Our method uses a conditional U-Net to generate unique perturbations within salient regions, addressing the limitations of existing DOV methods designed for natural images when extended to medical imaging.

    \item We introduce Laplacian regularization in the watermark generator to penalize high-frequency perturbations, encouraging the model to produce perturbations that are robust to downsampling.

    \item We perform extensive experiments on CXRs to demonstrate watermark effectiveness, imperceptibility, and transferability.
\end{itemize}

\section{Related Work} \label{sec:related_work}

\paragraph{Medical Data Protection}

The rapid expansion of artificial intelligence in healthcare has raised significant concerns regarding unauthorized exploitation of medical datasets~\cite{senbekov2020recent}. To address these concerns, medical image watermarking techniques have been developed to embed ownership information directly into medical images using methods such as discrete wavelet transform (DWT) and device fingerprinting~\cite{poonam2022advances,kancharla2024medical}. However, traditional watermarking may not be suitable for dataset ownership verification since embedded watermarks can hardly be inherited by trained models. More recently, Sun~et~al.~\cite{sun2024salm} proposed the Sparsity-Aware Local Masking (SALM) method, which generates unlearnable examples specifically designed for medical data by selectively perturbing significant pixel regions, leveraging the inherent sparse nature of medical images while maintaining clinical utility. However, similar to other unlearnable example methods~\cite{huang2021unlearnable}, Sun~et~al. requires watermarking over 90\% of the training data to ensure efficacy, which  would limit its practical applicability.

\paragraph{Data Ownership Protection (Auditing) with Watermark}

Dataset Ownership Verification (DOV) has emerged as a promising solution to protect dataset copyrights in deep learning~\cite{li2023blackbox,maini2021dataset}. Existing DOV methods can be categorized into two main types based on their reliance on internal features (IF) versus external features (EF)~\cite{shao2025databench}. IF-based methods exploit inherent data characteristics such as model overfitting through membership inference~\cite{maini2021dataset}. EF-based methods introduce artificial features into datasets, primarily through backdoor watermarking~\cite{adi2018turning,li2022untargeted,tang2023cleanLabel} that induces identifiable model behaviors, or non-poisoning watermarks~\cite{sablayrolles2020radioactive,guo2024domain} that embed features without causing misclassification. For unlabeled data, watermarking has been extended to self-supervised learning scenarios to protect pre-trained encoders~\cite{wu2022watermarking,cong2022sslguard}. However, recent adversarial evaluations~\cite{zhu2025evading,shao2025databench} have revealed that current DOV methods remain vulnerable to evasion attacks, underscoring the need for more robust dataset auditing approaches. 

\section{Conditional Scale-Invariant Clean-label Backdoor Watermark}

\begin{figure*}[!t]
    \centering
    \includegraphics[width=\linewidth]{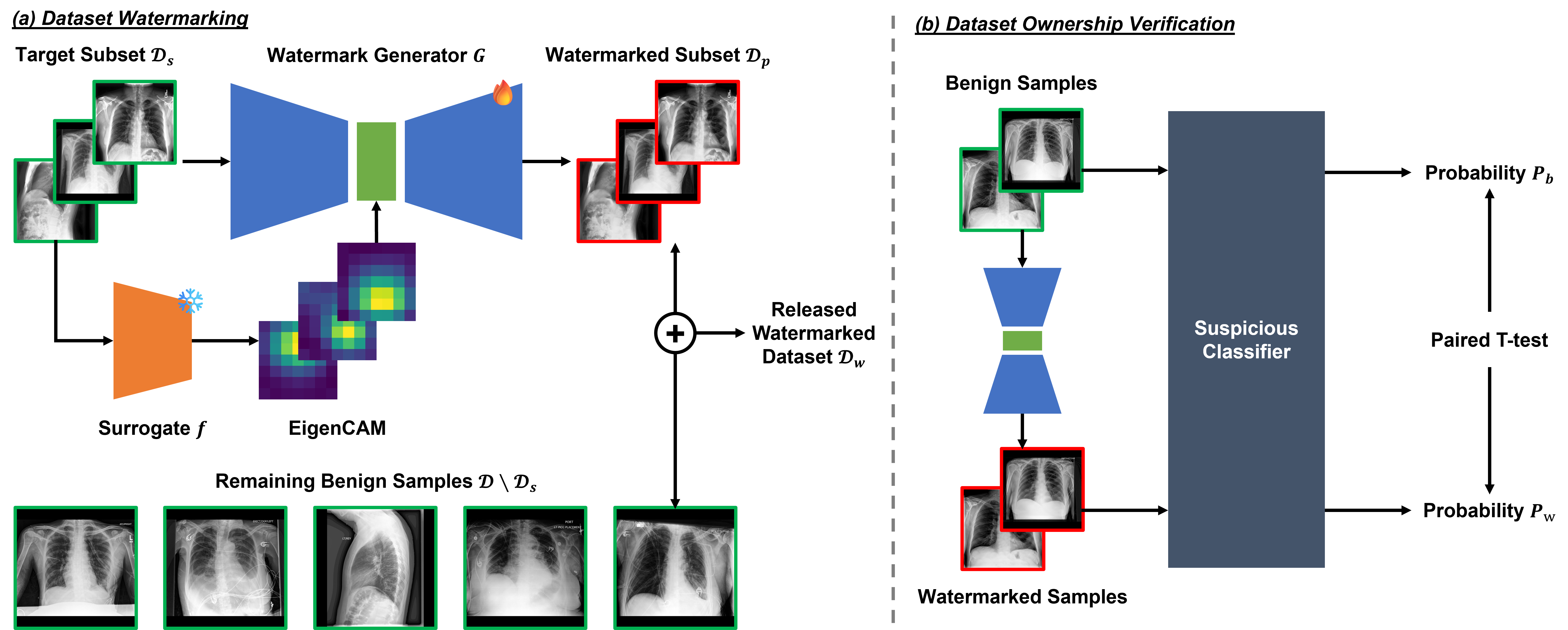}
    \caption{The main pipeline of X-Mark. First, a conditional U-Net is trained to generate sample-specific watermarks within salient regions of the medical image. Second, the watermarked dataset is created by embedding watermarks within a subset of target class samples and combining them with the remaining dataset. Finally, black-box dataset ownership verification is performed using hypothesis testing to detect whether watermarked, non-target class samples were misclassified by the suspicious model.}
    \label{fig:overview}
\end{figure*}

\subsection{Preliminaries}

\paragraph{Threat model} 

We focus on dataset ownership verification with watermark in medical image classification. There are two parties involved: a dataset owner and an unauthorized user. The dataset owners release their dataset for research or other innocent purposes, while protecting its copyright and preventing unauthorized use. The unauthorized user has incentive to train their commercial model without permission from the dataset owner. If the owner finds a suspicious model, unauthorized use of watermarked dataset can be verified by detecting characteristic behaviors in a black-box setting.

\paragraph{Main Pipeline of Dataset Watermarking} 

Consider a benign dataset $\mathcal{D} = \{(\mathbf{x}_i, \mathbf{y}_i\}^{N}_{i=1}$ containing $N$ samples. Each sample includes a medical image $\mathbf{x}_i \in \mathbb{R}^{H \times W}$ and its labels $\mathbf{y}_i \in \{0,1\}^K$. Specifically, this formulation describes a multi-label setting, where each $\mathbf{y}^{(k)}$ is a binary label for which $\mathbf{y}^{(k)}_{i} = 1$ indicates the presence of $k$-th pathology and vice versa. In contrast to multi-class settings, each sample may contain multiple pathologies.

The dataset owner selects a subset $\mathcal{D}_s \subset \mathcal{D}$ of samples from their dataset with watermarking rate $\gamma$. Generally, these samples share the same target label $y_t$ defined under a multi-class setting as $y_t \in \{1,2,...,K\}$. This formulation can be extended to the multi-label setting by selecting a specific target label $y^{(k)}$. Here, the target and non-target classes are defined as $y^{(k)}_t = 1$ and $y^{(k)}_{nt} = 0$ respectfively. For each sample $\mathbf{x}_i \in \mathcal{D}_s$, its watermarked version $\hat{\mathbf{x}}_i = G(\mathbf{x}_i;\theta)$ is generated using a watermark generator. This set of watermarked samples $\mathcal{D}_p$ is then combined with the remaining dataset $\mathcal{D} \setminus \mathcal{D}_s$ to create the watermarked dataset $\mathcal{D}_w$, and is subsequently released for legitimate use. 

Consider a suspicious model in a black-box setting, where the owner only has access to the model's API. The dataset owner intends to determine whether this model was trained on the watermarked dataset $\mathcal{D}_w$. A hypothesis-test-guided method based on the model's predicted probabilities on watermarked verification samples generated by $G(\cdot;\theta$) are used to determine unauthorized use of dataset. Specifically, this looks for characteristic backdoor behaviors where the suspicious model will behave normally on benign samples but misclassify the watermarked samples.

\subsection{Proposed Method}

As described in previous sections, existing DOV methods designed for natural images are incompatible with the constraints imposed by medical imaging. To overcome these limitations and ensure watermark stealthiness and robustness, we propose X-Mark, a sample-specific clean-label backdoor watermarking method for medical image copyright protection. Our work builds upon recent advances in clean-label backdoor watermarks, where a self-supervised, specialized watermark generator is used to embed watermark samples into the benign dataset \cite{wang2025sscl}. We condition the watermark generator to produces perturbations within salient regions, and introduce Laplacian regularization to achieve watermark scale-invariance. As shown in Figure \ref{fig:overview}, our method consists of three steps: (1) Training the conditional scale-invariant watermark generator, (2) Dataset watermarking, and (3) Dataset ownership verification.

\paragraph{Conditional Scale-Invariant Watermark Generator}

The goal of the watermark generator is to ensure that a suspicious model $f_w(\cdot;\omega)$ would learn the association between watermarked sample and the presence of target label $k$. Specifically, the suspicious model $f_w(\cdot;\omega)$ should trigger the backdoor when watermark signal (\textit{i.e.,} perturbations) present and cause misclassifications to  $y^{(k)}_t$. This presents a dual objective that the samples produced by the watermark generator $G(\cdot;\theta)$ must simultaneously satisfy in the clean-label setting:

\noindent\textbf{Objective I.}
During dataset watermarking, only samples from the target class are watermarked in the released dataset. The watermark generator should produce watermarked samples from the target class containing perturbations that reinforce the watermark-label association, such that when these samples would be misclassified by a suspicious classifier $f_w(\cdot;\omega)$.
\begin{equation}
    f_w(G(\mathbf{x}_{t,i}; \theta); \omega)^{(k)} = 0
\end{equation}

\noindent\textbf{Objective II.} During DOV, only samples from the non-target class are used for verification. The watermark generator should produce watermarked samples from the non-target class such that the suspicious classifier $f_w(\cdot;\omega)$ would misclassify them as $y^{(k)}_t$.
\begin{equation}
    f_w(G(\mathbf{x}_{nt,i}; \theta); \omega)^{(k)} = 1
\end{equation}
    
We use a conditional U-Net autoencoder as our watermark generator $G(\cdot;\theta)$. Unlike SSCL-BW \cite{wang2025sscl}, which used the original U-Net architecture \cite{ronneberger2015u}, our model adopts a residual U-Net with strided convolutions to improve computational efficiency at higher image resolutions. This reduces parameter count by $25\%$ and MAC operations by $72\%$ compared to the original U-Net. Additionally, we incorporate an EigenCAM-based \cite{muhammad2020eigen} saliency conditioning module at the bottleneck via convolutional projection to condition the generator to produce perturbations within salient regions of the sample. We choose EigenCAM over other gradient-based activation maps (e.g., Grad-CAM) for its efficiency.

To train the watermark generator $G(\cdot;\theta)$ such that watermarked samples satisfy the dual objective and medical imaging constraint, we use a surrogate benign classifier $f(\cdot;\omega)$ and training objective consisting of four components: (1) Target sample loss $\mathcal{L}_{t}$, (2) Non-target sample loss $\mathcal{L}_{nt}$, (3) Perceptual similarity loss $\mathcal{L}_{\text{LPIPS}}$ and (4) Laplacian pyramid loss $\mathcal{L}_{\text{Lap}}$. While the first three components are inspired by prior work \cite{wang2025sscl}, we introduce $\mathcal{L}_{\text{Lap}}$ to regularize the watermark generator by penalizing high-frequency perturbations and mitigating strong, unrealistic perturbations, encouraging the model to produce perturbations that are robust to downsampling. The overall training objective is the weighted average of all four components:
\begin{equation}
    \mathcal{L} = \lambda_{t} \mathcal{L}_{t} + \lambda_{nt}  \mathcal{L}_{nt} + \lambda_{\text{LPIPS}} \mathcal{L}_{\text{LPIPS}}  + \lambda_{\text{Lap}} \mathcal{L}_{\text{Lap}}
\end{equation}
where $\lambda_{t} = \lambda_{nt} = 1$, $\lambda_{\text{Lap}} = 3$, and $\lambda_{\text{LPIPS}} = 10$.

The target sample loss $\mathcal{L}_{t}$ measures the loss of the surrogate model on watermarked samples from the target class:
\begin{equation}
    \mathcal{L}_{t} = \mathbb{E}_{\mathbf{x}_{t} \sim \mathcal{D}_{t}} \mathcal{H}_b(f(G(\mathbf{x}_{t}; \theta); \omega)^{(k)}, 0)
\end{equation}
where $\mathcal{H}_b$ is binary cross-entropy (BCE) loss, $\mathcal{D}_t \subset \mathcal{D}$ is the set of target samples.

Similarly, the non-target sample loss $\mathcal{L}_{nt}$ measures the loss of the surrogate model on watermarked samples from the non-target class:
\begin{equation}
    \mathcal{L}_{nt} = \mathbb{E}_{\mathbf{x}_{nt} \sim \mathcal{D}_{nt}} \mathcal{H}_b(f(G(\mathbf{x}_{nt}; \theta); \omega)^{(k)}, 1)
\end{equation}
where $\mathcal{D}_{nt} \subset \mathcal{D}$ is the set of non-target samples.

To ensure watermark imperceptibility and preserve diagnostic quality, the LPIPS metric measures the perceptual similarity between benign and watermarked samples from both target and non-target classes \cite{zhang2018unreasonable}:
\begin{equation}
    \mathcal{L}_{\text{LPIPS}} = \mathbb{E}_{\mathbf{x} \sim \mathcal{D}}\ \text{LPIPS}(G(\mathbf{x};\theta), \mathbf{x})
\end{equation}

To achieve watermark scale-invariance, we incorporate Laplacian regularization using the Laplacian pyramid loss $\mathcal{L}_{\text{Lap}}$ to penalizes differences across multiple frequency levels \cite{denton2015deep,bojanowski2017optimizing}:
\begin{equation}
    \mathcal{L}_{\text{Lap}} = \sum^{L}_{l=1} \mathbb{E}_{\mathbf{x} \sim \mathcal{D}} \lVert \mathcal{L}^{(l)}(G(\mathbf{x}; \theta)) -  \mathcal{L}^{(l)}(\mathbf{x})\rVert_1
\end{equation}
where $\mathcal{L}^{(l)}$ is the $l$-th level of the Laplacian pyramid decomposition and $L=3$ is the number of pyramid levels.

We constraint the perturbation between the benign sample $\mathbf{x}_i$ and its watermarked version $\hat{\mathbf{x}}_t$ by the budget $\epsilon$:
\begin{equation}
    \lVert G(\mathbf{x}_i; \theta) - \mathbf{x}_i \rVert_\infty \leq \epsilon
\end{equation}
Following prior work, we set $\epsilon = 16/255$ \cite{li2022untargeted}.

\paragraph{Dataset Watermarking}

We embed watermarked samples into the benign dataset following prior clean-label backdoor watermarks \cite{li2022untargeted}. We select a subset $\mathcal{D}_s \subset \mathcal{D}$ of benign samples from the target class with watermarking rate $\gamma$. For each sample $\mathbf{x}_{t,i}$, we generate its watermarked version $\hat{\mathbf{x}}_{t,i}$ using the watermark generator $G(\cdot;\theta)$. These watermarked samples are then merged with the remaining dataset $\mathcal{D} \setminus \mathcal{D}_s$ to create the watermarked dataset $\mathcal{D}_w$, and is subsequently released for legitimate use. The goal is that a suspicious model trained on the watermarked dataset would inherit the watermark-label association and trigger the backdoor upon verification. Furthermore, our watermarks are hard to detect with manual inspection due to the sample-specific clean-label nature of our method.

\paragraph{Dataset Ownership Verification}

Suppose an unauthorized user fine-tunes a pre-trained model $f(\cdot;\omega)$ on the watermarked dataset $\mathcal{D}_w$, resulting in the backdoored model $f_w(\cdot;\omega)$. Given this suspicious model in black-box setting, we follow prior clean-label backdoor watermarks for probability-available verification \cite{li2022untargeted}. Specifically, if the model was trained on the watermarked dataset $\mathcal{D}_w$, the watermark should trigger the backdoor and verification samples would be misclassified as $y^{(k)}_t = 1$. To prevent selection bias, probability-available verification uses a hypothesis-test to check if the posterior probability for the target class of benign samples is significantly lower than that of their watermarked versions. 

Formally, let $\mathcal{X}$ denote the set of benign samples from the non-target class, and $\mathcal{X}' = G(\mathcal{X};\theta)$ be their watermarked versions. Let $P_b=f_w(\mathcal{X})^{(k)}$ and $P_v=f_w(\mathcal{X}')^{(k)}$ denote the posterior probabilities of $\mathcal{X}$ and $\mathcal{X}'$ on the target class $y^{(k)}_{t}$ respectively. Consider the null hypothesis:
\begin{align}
    H_o : P_b + \tau = P_v \\ H_1: P_b + \tau < P_v
\end{align}
where $\tau \in [0,1]$ is the accepted margin. If the resulting p-value of one-sided paired t-test is significant, $H_0$ is rejected and we conclude that the suspicious model was trained on the watermarked dataset $\mathcal{D}_w$. Additionally, we compute the confidence score $\Delta P = P_b - P_v$, where a larger $\Delta P$ indicates higher verification effectiveness. Statistical significance is defined as $p<0.05$.

\section{Experiments}

\subsection{Experimental Setup}

\paragraph{Dataset}

We conduct experiments on CheXpert \cite{irvin2019chexpert}, containing $n=224,316$ frontal and lateral CXRs from $65,240$ patients annotated for the presence of 14 radiological findings. Specifically, we focus on the six CheXpert-defined competition labels: Atelectasis, Cardiomegaly, Consolidation, Edema, and Pleural Effusion as well as No Finding, which represents the absence of any pathology.

\paragraph{Implementation}

We train the surrogate model $f(\cdot;\omega)$ on the benign dataset using an ImageNet pre-trained ResNet18 for 5 epochs with AdamW optimizer, batch size of 32, and learning rate of 1e-4 at the resolution of $224\times224$. To train the watermark generator $G(\cdot;\theta)$, we select $10\%$ of samples from the target class and an equal number of samples from the non-target class to ensure a balanced training set. During generator training, we compute EigenCAMs from the final convolutional layer of the surrogate model and use them as conditioning input to guide perturbations towards salient regions. The generator is optimized using AdamW with a batch size of 32 and learning rate of 1e-4 for up to 100 epochs at resolution $1024\times1024$. Finally, the backdoored model $f_w(\cdot;\omega)$ is trained similarly to the surrogate model using an ImageNet pre-trained ResNet18 for 5 epochs with AdamW optimizer, batch size of 32, and learning rate of 1e-4 at the resolution of $224\times224$. All models are implemented in PyTorch and trained with random augmentations, mixed-precision, and 8 NVIDIA L40S GPUs on a shared HPC.

\paragraph{Evaluation Metrics}

\begin{table*}[!t]
    \centering
    \caption{Benign accuracy (\%), watermark success rate (\%), and perceptual similarity of dataset watermarking on CheXpert.}
    \label{tab:baseline_results}
    \small
    \begin{tabular}{c|c|c|c|c} \toprule \rowcolor{gray!25}
        Type$\downarrow$ & Method$\downarrow$, Metric$\rightarrow$ & BA (\%) & WSR (\%) & LPIPS \\ \hline
        Benign & No Attack & 81.89 & - & - \\ \hline
        \multirow{4}{*}{Poison-Label} & BadNets \cite{gu2017badnets} & 80.69 & 100 & 0.009 \\
        & Blended \cite{chen2017targeted} & 83.23 & 97.31 & 0.003 \\
        & WaNet \cite{nguyen2021wanet} & 82.48 & 16.81 & 0.004 \\
        & UBW-P \cite{li2022untargeted} & 84.28 & 22.45 & 0.009 \\ \hline
        \multirow{3}{*}{Clean-Label} & Label-Consistent \cite{turner2019label} & 80.69 & 100 & 0.009 \\
        & SSCL-BW \cite{wang2025sscl} & $82.93$ & $100$ & $0.023$ \\
        & X-Mark (Ours) & $84.88$ & $100$ & $0.020$ \\ \hline
    \end{tabular}
\end{table*}

\begin{table*}[!t]
    \centering
    \caption{Effectiveness ($\Delta P$ and p-value) of probability-available dataset ownership verification on CheXpert.}
    \label{tab:ttest_results}
    \small
    \begin{tabular}{c|c|c|c|c} \toprule \rowcolor{gray!25}
        Method$\downarrow$ & Metric$\downarrow$, Scenario$\rightarrow$ & Ind-W & Ind-M & Malicious \\ \hline
        \multirow{2}{*}{SSCL-BW \cite{wang2025sscl}} & $\Delta P$ & $-0.0004$ & $0.2925$ & $0.8705$ \\
        & p-value & $1.0$ & \cellcolor{red!25}$0.0580$ & $10^{-60}$ \\ \hline
        \multirow{2}{*}{X-Mark (Ours)} & $\Delta P$ & $-0.0005$ & $0.2564$ & $0.8378$ \\
        & p-value & $1.0$ & \cellcolor{green!25} $0.4102$ & $10^{-53}$ \\ \hline
    \end{tabular}
\end{table*}

We measure the performance of our watermarking method across dataset watermarking and DOV tasks. For dataset watermarking, we evaluate watermark effectiveness and imperceptibility using benign accuracy (BA), watermark success rate (WSR) and perceptual similarity (LPIPS). Briefly, BA measures the accuracy of benign samples being correctly classified, while WSR measures the proportion of non-target watermarked samples that are misclassified as the target-class by the suspicious model. While AUROC is primarily used as the evaluation metric in medical image classification, we compute the optimal F1 threshold on the validation set and calculate accuracy on the test set to maintain consistency with prior works. For DOV, we evaluate verification effectiveness using confidence $\Delta P \in [-1, 1]$ and p-value $p\in[0,1]$. A larger $\Delta P$ and smaller $p$ indicates a high likelihood that the suspicious model was trained on the watermarked dataset.

\subsection{Performance of Dataset Watermarking}

\paragraph{Setting}

We evaluate the effectiveness and imperceptibility by comparing our method with existing backdoor watermarking methods. For the poison-label setting, we include BadNets \cite{gu2017badnets}, Blended attack \cite{chen2017targeted}, WaNet \cite{nguyen2021wanet}, and UBW-P \cite{li2022untargeted} as baselines. For the clean-label setting, we include Label-Consistent \cite{turner2019label} and SSCL-BW \cite{wang2025sscl} as baselines. Additionally, we include the ``No Attack" scenario as our BA reference. The target label is set as ``No Finding" and watermarking rate is set as $\gamma = 0.1$.

\paragraph{Results}

\begin{figure}[!t]
    \centering
    \includegraphics[width=0.8\linewidth]{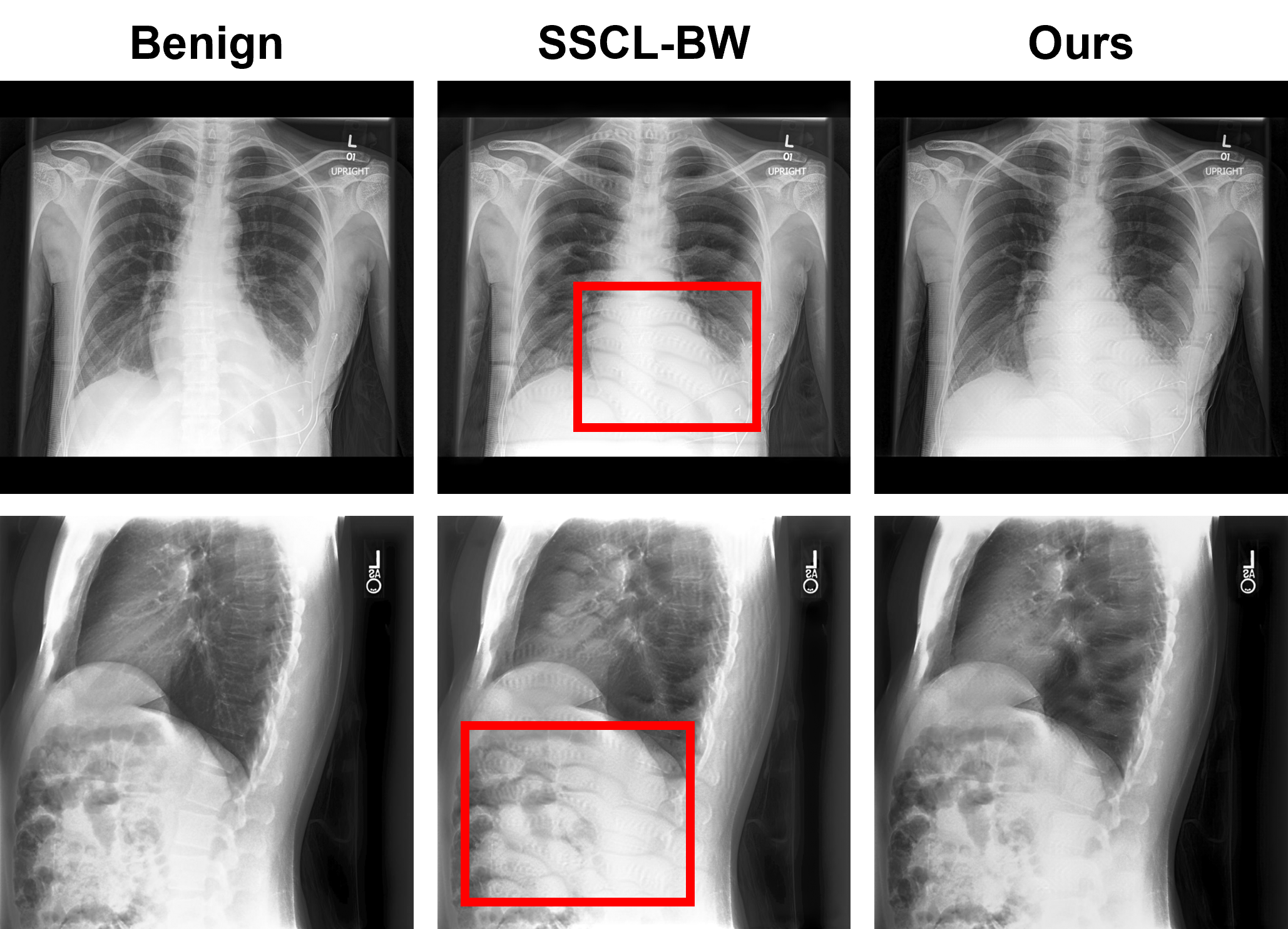}
    \caption{Example watermarked samples from SSCL-BW and X-Mark. Red box indicates region of strong perturbations, resulting in anatomically improbable structures that are easy to detect upon manual inspection. Saliency conditioning limits perturbations within salient regions (chest) while Laplacian regularization mitigates strong, unrealistic perturbations.}
    \label{fig:examples}
\end{figure}

As shown in Table \ref{tab:baseline_results}, our method demonstrates strong watermark effectiveness while maintaining imperceptibility, achieving highest BA of $84.88\%$ and WSR of $100\%$. Among the poison-label baselines, BadNets and Blended attacks achieve high WSRs ($100\%$ and $97.31\%$). However, they rely on static trigger patterns that need to large enough to survive downsampling, making them easily detectable. WaNet watermarks fail to survive downsampling, only achieving a WSR of $16.81\%$. Similarly, UBW-P with a WSR of $22.45\%$ scales poorly to the multi-label setting due to its untargeted nature. In the clean-label setting, Label-Consistent and SSCL-BW both reach WSR of $100\%$. Label-Consistent relies on static triggers, while SSCL-BW generates strong perturbations (LPIPS of 0.023) that produce unrealistic anatomy, as shown in Figure \ref{fig:examples}. Results indicate that CXR datasets are well-suited for dataset watermarking, likely due to their limited visual diversity compared to natural images, allowing perturbations to remain in-distribution and exploit shortcut learning \cite{jabbour2020deep}.

\subsection{Performance of Dataset Ownership Verification}

\paragraph{Setting} 

Following prior work \cite{li2022untargeted,li2023blackbox}, we evaluate the effectiveness of probability-available verification across three scenarios: (1) Independent Watermark (Ind-W), where the suspicious model trained on the watermarked dataset is queried with a different watermark than that produced by our method. (2) Independent Model (Ind-M), where a suspicious benign model is queried with the watermarks produced by our method. (3) Malicious, where the suspicious model trained on the watermarked dataset is queried with the watermarks produced by our method. Additionally, we also report the verification effectiveness of the SSCL-BW \cite{wang2025sscl} as it is our primary baseline. We randomly sample $n=100$ benign samples from the dataset and set $\tau=0.25$ for the hypothesis test.

\begin{figure}[!t]
    \centering
    \includegraphics[width=0.8\linewidth]{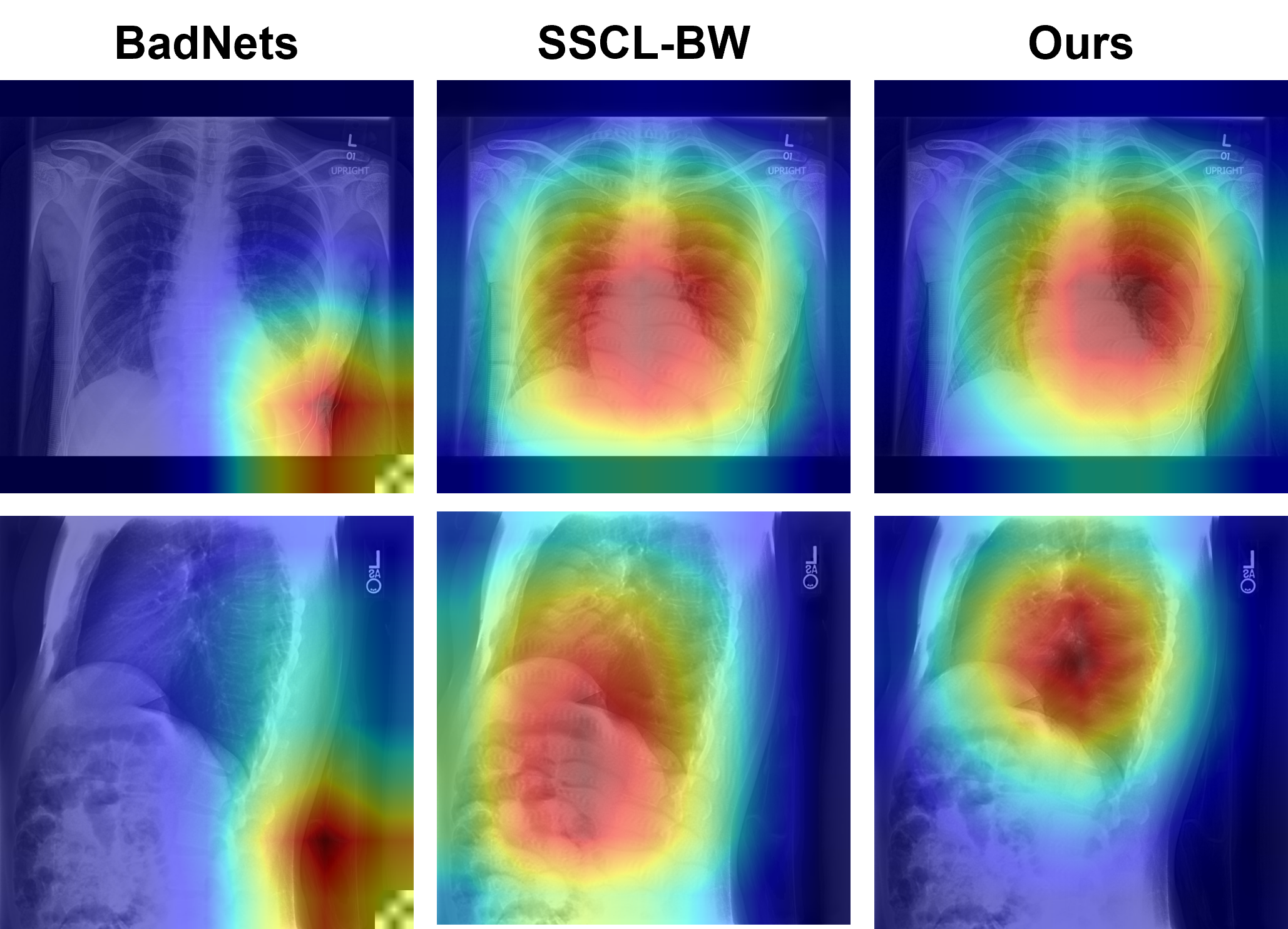}
    \caption{Watermarked samples and their EigenCAM-based saliency maps from backdoored models using BadNets, SSCL-BW, and X-Mark. For BadNets, the EigenCAM mainly focuses on the trigger, while focusing on regions with the largest perturbations in SSCL-BW. In contrast, in our method, the EigenCAM focuses on the salient regions (i.e., the chest), making the backdoor difficult to detect even with automated methods.}
    \label{fig:saliency}
\end{figure}

\paragraph{Results} 

As shown in Table \ref{tab:ttest_results}, our method demonstrates effective verification in the probability-available black-box setting. It accurately detects whether the suspicious model $f_w(\cdot;\omega)$ was trained on the watermarked dataset $\mathcal{D}_w$ in the ``Malicious" scenario, achieving $\Delta P > 0.8$ and $p<0.001$. It also does not produce false positives in the ``Ind-W" scenario, where $\Delta P \approx 0$ and $p=1.0$. However, in the ``Ind-M" scenario, we observe $\Delta P=0.26$ and $p=0.41$, indicating a greater probability of false positives for benign models. While SSCL-BW exhibits comparable performance in the ``Malicious" and ``Ind-W" scenarios, its performance in ``Ind-M" in substantially worse with $p=0.06$. We attribute this behavior to the challenges discussed previously in medical imaging. It is likely that watermark perturbations are in-domain to the feature representation naturally learned by classifier, causing benign models to also exhibit backdoored behavior. A potential solution may involve injecting out-of-distribution perturbations \cite{guo2024domain,feng2022fiba}. 

\subsection{Transferability of X-Mark}

We explore the transferability and robustness of our watermarking method across two settings: (1) Watermark scale-invariance, and (2) Model-agnostic transferability.

\paragraph{Watermark Scale-Invariance}

\begin{figure}[!t]
    \centering
    \includegraphics[width=\linewidth]{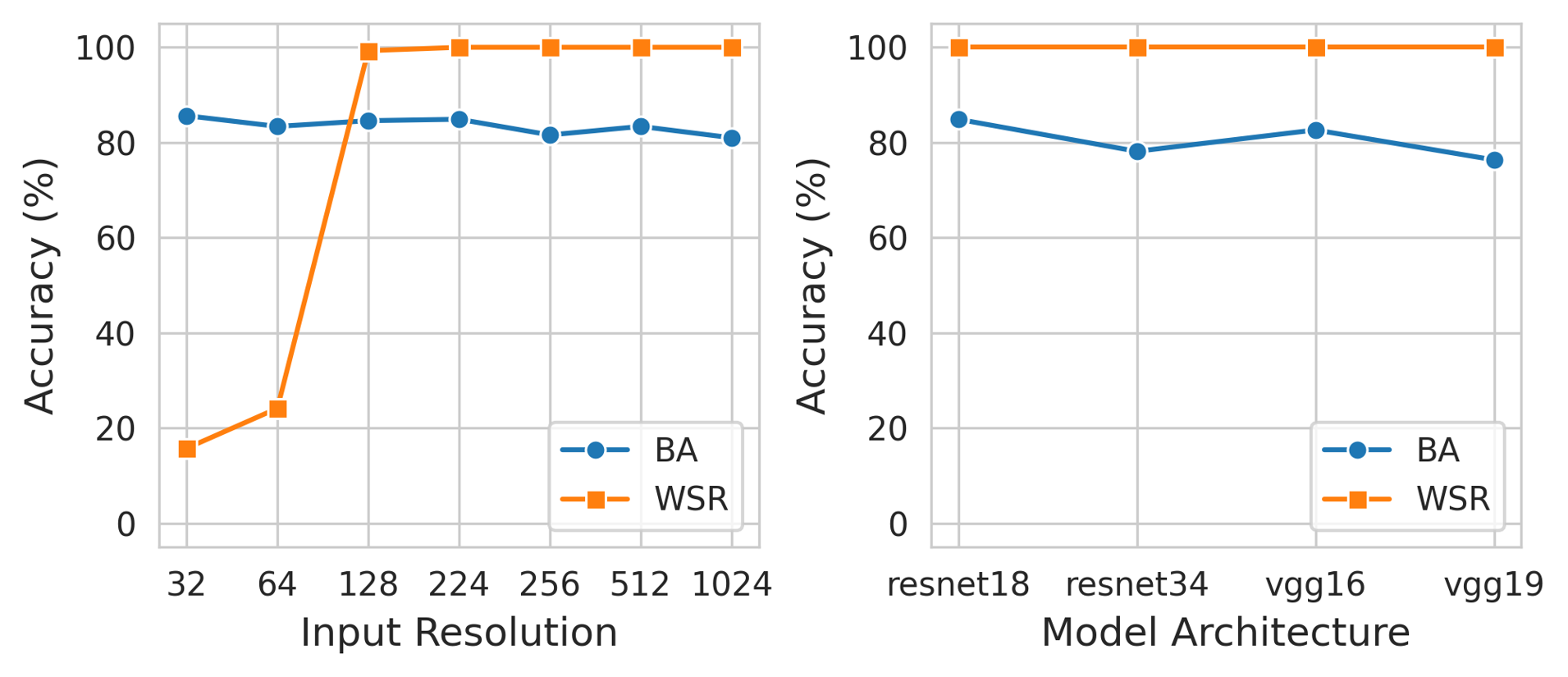}
    \caption{Transferability of X-Mark, demonstrating watermark scale-invariance and model-agnostic transferability.}
    \label{fig:transfer_results}
\end{figure}

We train backdoored models $f_w(\cdot;\omega)$ on the watermarked dataset $\mathcal{D}_w$ at resolutions of $32\times32$, $64\times64$, $128\times128$, $224\times224$, $256\times256$, $512\times512$ and $1024\times1024$, using the same settings described previously. All images are resized to the target resolution using bilinear interpolation while preserving aspect ratio via zero-padding. As shown in Figure \ref{fig:transfer_results}, the backdoored models inherit the watermark-label association even when images are downsampled. We observe that consistent BA and WSR up to $128\times128$ resolution. However, watermark effectiveness decreases when the resolution is further reduced to $64\times64$ and $32\times32$. This demonstrates that the watermark generator produces perturbations that are robust to downsampling and remain scale-invariant up to a certain resolution.

\paragraph{Model-Agnostic Transferability}

We train backdoored models $f_w(\cdot;\omega)$ using the watermarked dataset $\mathcal{D}_w$ using four distinct architectures: ResNet18, ResNet34, VGG16-BN, and VGG19-BN. All models are trained using the same settings described previously. As shown in Figure \ref{fig:transfer_results}, all architectures inherit the backdoor behavior, consistently reaching a WSR of $100\%$. This indicates that the watermarked samples are model-agnostic and retain their effectiveness across different model architectures.

\subsection{Ablation Study}

We explore dataset watermarking performance of our method under various settings, specifically the impact of the proposed components (saliency conditioning and Laplacian regularization) and core hyperparameters (watermarking rate $\gamma$ and perturbation budget $\epsilon$). All experiments are conducted under the same settings described previously.

\paragraph{Impact of Saliency Conditioning}

\begin{figure}[!t]
    \centering
    \includegraphics[width=\linewidth]{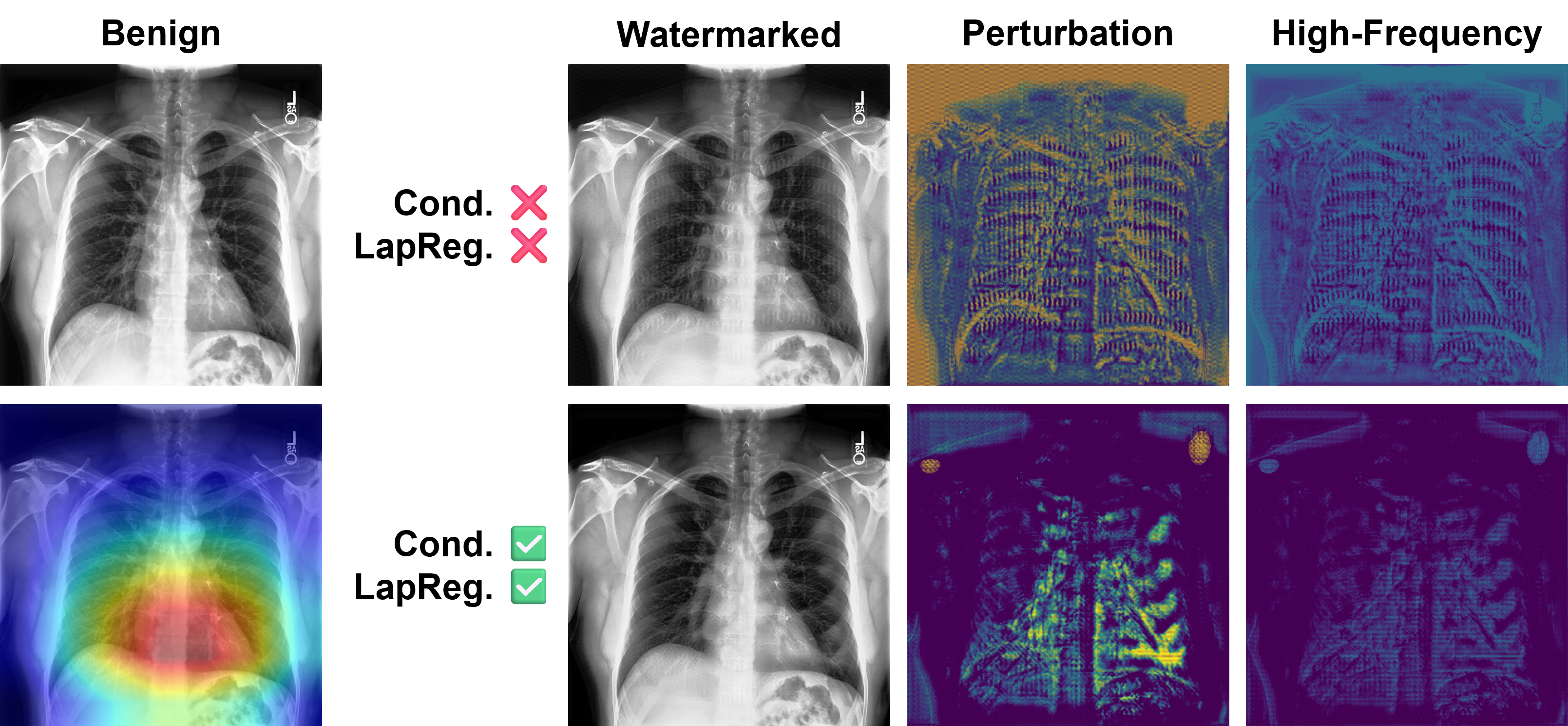}
    \caption{An example illustrating the impact of EigenCAM-based saliency conditioning and Laplacian regularization.}
    \label{fig:component_example}
\end{figure}

To study the impact of saliency conditioning, we train watermark generators $G(\cdot;\theta)$ with and without the EigenCAM-based saliency conditioning module. As shown in Table \ref{tab:component_ablation_results}, dataset watermarking performance is not affected by saliency conditioning, with comparable BA, WSR, and LPIPS metrics across both settings. We observe a decrease in $\mathcal{L}_{\text{Lap}}$ even when Laplacian regularization is not applied, likely due to the perturbations being concentrated within salient regions of the image. This behavior is visualized in Figure \ref{fig:component_example}.

\paragraph{Impact of Laplacian Regularization}

To study the impact of Laplacian regularization, we train watermark generators $G(\cdot;\theta)$ with and without the Laplacian pyramid loss $\mathcal{L}_{\text{Lap}}$ in the training objective. As shown in Table \ref{tab:component_ablation_results}, we observe an increase in BA ($\sim4\%$) and decrease in $\mathcal{L}_{\text{Lap}}$, while WSR and LPIPS remain unchanged. This indicates that Laplacian regularization mitigates high-frequency perturbations. We visualize the high-frequency information using the Laplacian filter $\nabla^2$ (i.e., high-pass filter) in Figure \ref{fig:component_example}.

\begin{table}[!t]
    \centering
    \caption{Impact of EigenCAM-based saliency conditioning and Laplacian regularization of dataset watermarking performance on CheXpert.}
    \label{tab:component_ablation_results}
    \small
    \begin{tabular}{c|c|c|c|c|c} \toprule \rowcolor{gray!25}
        Cond. & LapReg. & BA (\%) & WSR (\%) & LPIPS & $\mathcal{L}_{\text{Lap}}$ \\ \hline
        \xmark & \xmark & $80.24$ & $100$ & $0.019$ & $0.042$ \\
        \xmark & \cmark & $84.58$ & $100$ & $0.019$ & $0.014$ \\
        \cmark & \xmark & $80.39$ & $100$ & $0.020$ & $0.019$ \\ \rowcolor{green!25}
        \cmark & \cmark & $84.88$ & $100$ & $0.020$ & $0.013$ \\ \hline
    \end{tabular}
\end{table}

\paragraph{Impact of Watermarking Rate}

\begin{figure}[!t]
    \centering
    \includegraphics[width=\linewidth]{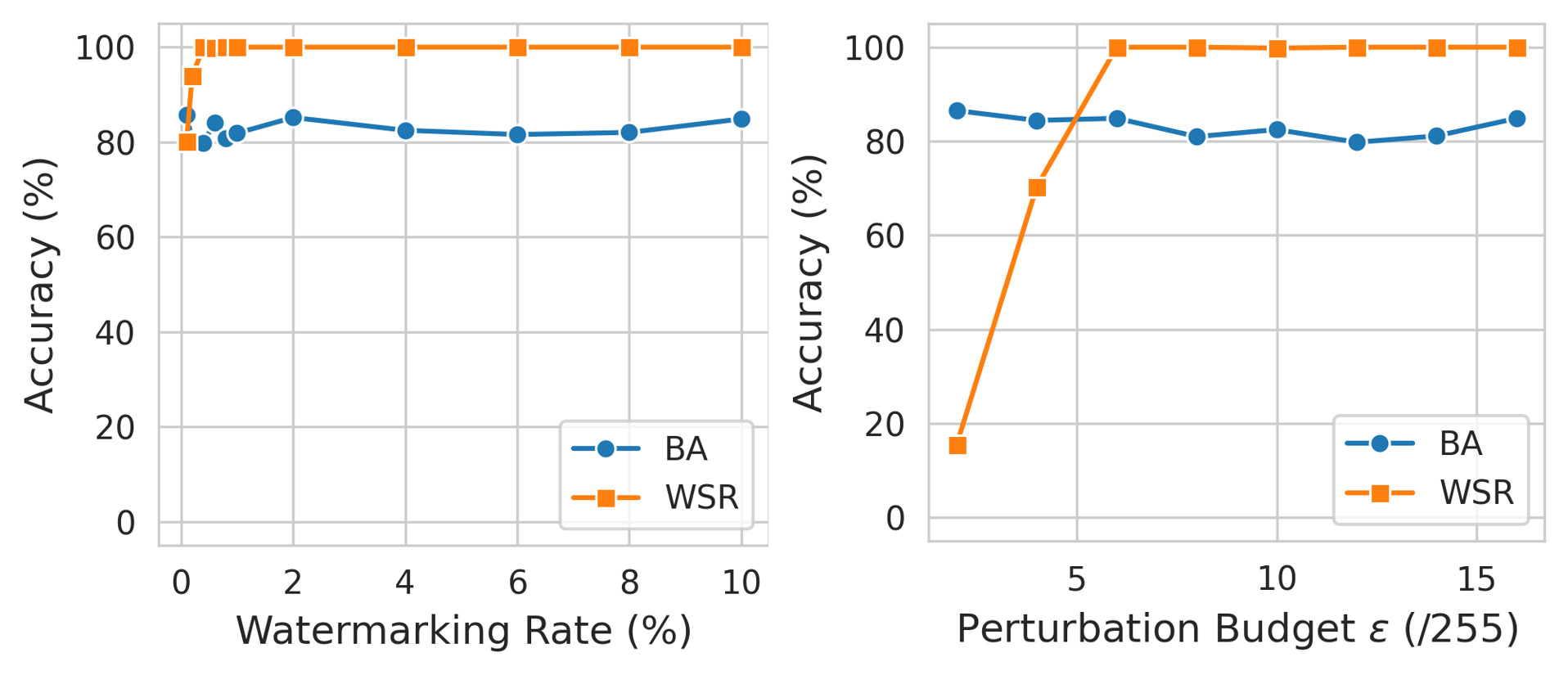}
    \caption{Impact of $\gamma$ and $\epsilon$ on X-Mark dataset watermarking.}
    \label{fig:watermark_ablation_results}
\end{figure}

To study the impact of watermarking rate $\gamma$, we train backdoored models $f_w(\cdot;\omega)$ with $\gamma$ ranging from $0.1\%$ to $10\%$. As shown in Figure \ref{fig:watermark_ablation_results}, even at a low rate of $\gamma = 1\%$, the WSR reaches $100\%$ while BA is unaffected. This indicates that increasing $\gamma$ improves watermark effectiveness without impacting model performance on benign samples. The high WSR observed at very low $\gamma=0.01\%$ can be attributed to the sensitivity of benign models to the watermark, consistent with the observations made in the ``Ind-M" scenario.

\paragraph{Impact of Perturbation Budget}

To study the impact of perturbation budget $\epsilon$, we train watermark generators $G(\cdot;\theta)$ with $\epsilon$ ranging from $2/255$ to $16/255$. As shown in Figure \ref{fig:watermark_ablation_results}, increasing $\epsilon$ leads to higher WSR, while BA remains comparable across all settings. Even at a budget of $\epsilon=6/255$, the WSR reaches $100\%$, demonstrating that effective watermarks can be achieved with relatively small perturbations. However, larger values of $\epsilon$ involve a trade-off with imperceptibility, as larger perturbations in watermarked samples would be evident upon manual inspection.

\subsection{Resistance to Adaptive Attacks}

We explore whether our method is resistant to existing backdoor defenses that an unauthorized user may employ to avoid detection. Using the same settings as previously described, we focus on two representative watermark removal attacks: (1) Model fine-tuning \cite{liu2017neural} and (2) Model pruning \cite{liu2018fine}.

\paragraph{Model Fine-tuning}

\begin{figure}[!t]
    \centering
    \includegraphics[width=\linewidth]{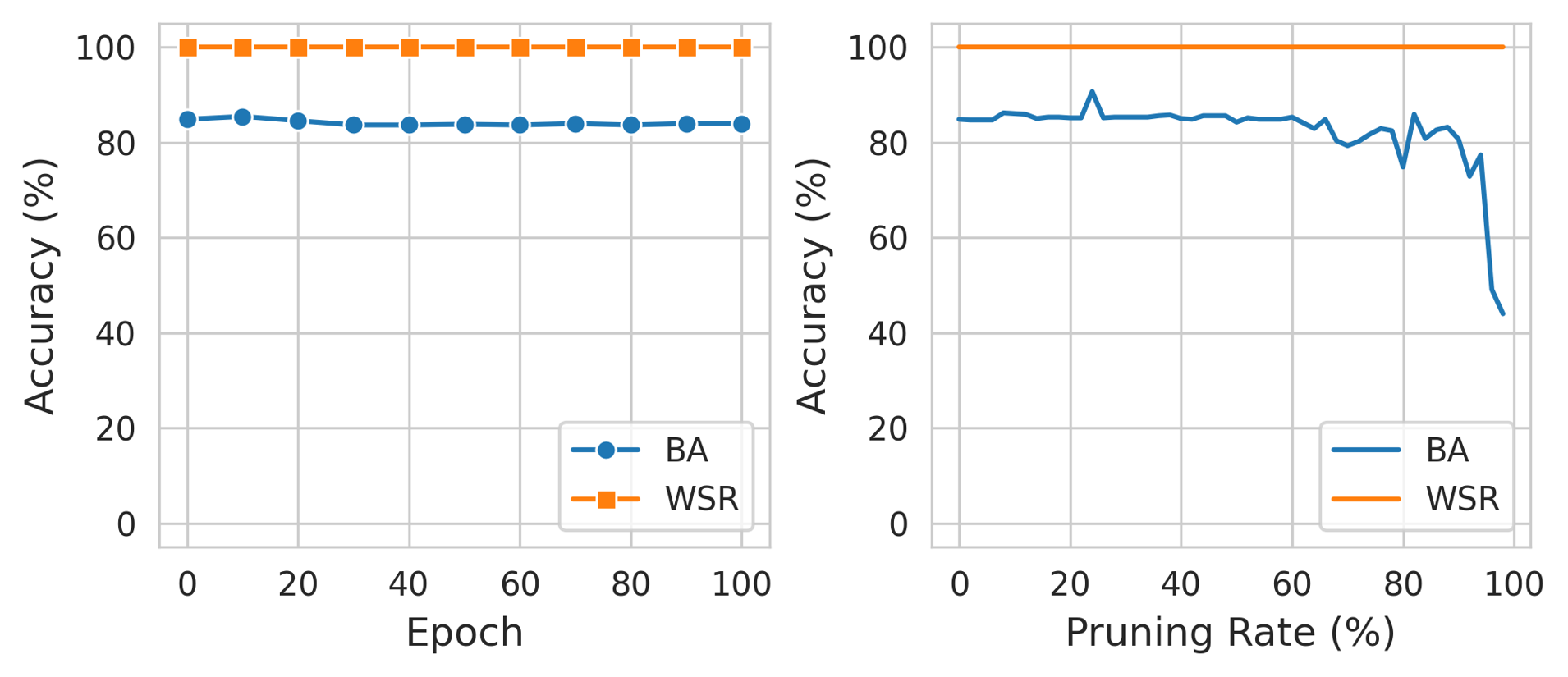}
    \caption{Resistance of X-Mark to model fine-tuning and pruning.}
    \label{fig:defense_results}
\end{figure}

We randomly select $10\%$ of benign samples from the dataset to fine-tune the backdoored model $f_w(\cdot:\omega)$ for up to 100 epochs. Specifically, only the fully-connected layers of $f_w(\cdot;\omega)$ are fine-tuned while convolulation layers are frozen. As shown in Figure \ref{fig:defense_results}, both BA and WSR generally remain consistent even after fine-tuning for 100 epochs. This indicates that model fine-tuning has little-to-no impact on watermark performance.

\paragraph{Model Pruning}

We randomly select $10\%$ of benign samples from the dataset to prune the latent representation of the backdoored model $f_w(\cdot:\omega)$  with a pruning rate $\beta \in \{0, 0.02, ..., 0.98\}$. Specifically, we use fine-pruning \cite{liu2018fine} on the output of the last convolutional layer. As shown in Figure \ref{fig:defense_results}, BA gradually decreases as pruning rate increases, while WSR remains unchanged across all rates. We attribute high WSR at high pruning rates due to watermarked samples consistently having higher prediction probabilities than benign samples, resulting in WSR of $100\%$.

\section{Conclusion}

In this paper, we propose X-Mark, a sample-specific clean-label backdoor watermarking method for copyright protection of CXR datasets. Our method addresses the challenges presented by CXRs, namely, the dynamic and high-resolution of scans, limited visual diversity, and need for preserving diagnostic quality. We use a U-Net watermark generator with an EigenCAM-based saliency conditioning module that embeds unique perturbations within salient regions of the CXR. Additionally, we introduce Laplacian regularization to penalize high-frequency perturbations and prevent strong, unrealistic perturbations, allowing the watermark to survive downsampling and be visually indistinguishable from benign samples. Experimental results on CheXpert show that X-Mark outperforms existing clean-label watermarking methods in effectiveness, imperceptibility, transferability, and resilience against watermark removal attacks. More importantly, it demonstrates robustness under medical imaging settings. Future work will explore extending our method to other medical imaging tasks (particularly, segmentation) and applications in medical VLMs.

\bibliographystyle{named}
\bibliography{ijcai26}

@article{senbekov2020recent,
  title={The recent progress and applications of digital technologies in healthcare: a review},
  author={Senbekov, Meiram and Saliev, Timur and Bukeyeva, Zarina and Almabayeva, Aigul and Zhanaliyeva, Murat and Aitenova, Nurgul and Toishibekov, Yerzhan and Fakhradiyev, Ildar and others},
  journal={International Journal of Telemedicine and Applications},
  volume={2020},
  year={2020},
  publisher={Hindawi}
}

@inproceedings{abadi2016deep,
  title={Deep learning with differential privacy},
  author={Abadi, Martin and Chu, Andy and Goodfellow, Ian and McMahan, H Brendan and Mironov, Ilya and Talwar, Kunal and Zhang, Li},
  booktitle={ACM SIGSAC Conference on Computer and Communications Security},
  pages={308--318},
  year={2016}
}

@inproceedings{huang2021unlearnable,
  title={Unlearnable examples: Making personal data unexploitable},
  author={Huang, Hanxun and Ma, Xingjun and Erfani, Sarah Monazam and Bailey, James and Wang, Yisen},
  booktitle={International Conference on Learning Representations},
  year={2021}
}

@article{sun2024salm,
  title={Medical Unlearnable Examples: Securing Medical Data from Unauthorized Training via Sparsity-Aware Local Masking},
  author={Sun, Weixiang and others},
  journal={arXiv preprint arXiv:2403.10573},
  year={2024}
}

@article{poonam2022advances,
  title={Advances in medical image watermarking: a state of the art review},
  author={Poonam and Arora, Sumit},
  journal={Multimedia Tools and Applications},
  volume={81},
  pages={1--35},
  year={2022},
  publisher={Springer}
}

@article{kancharla2024medical,
  title={Medical Image Data Provenance for Medical Cyber-Physical System},
  author={Kancharla, Pranay and others},
  journal={arXiv preprint arXiv:2403.15522},
  year={2024}
}

@article{li2023blackbox,
  title={Black-box Dataset Ownership Verification via Backdoor Watermarking},
  author={Li, Yiming and Zhu, Mingyan and Yang, Xue and Jiang, Yong and Wei, Tao and Xia, Shu-Tao},
  journal={IEEE Transactions on Information Forensics and Security},
  volume={18},
  pages={2318--2332},
  year={2023}
}

@inproceedings{maini2021dataset,
  title={Dataset Inference: Ownership Resolution in Machine Learning},
  author={Maini, Pratyush and Yaghini, Mohammad and Papernot, Nicolas},
  booktitle={International Conference on Learning Representations},
  year={2021}
}

@inproceedings{adi2018turning,
  title={Turning Your Weakness Into a Strength: Watermarking Deep Neural Networks by Backdooring},
  author={Adi, Yossi and Baum, Carsten and Cisse, Moustapha and Pinkas, Benny and Keshet, Joseph},
  booktitle={27th USENIX Security Symposium},
  pages={1615--1631},
  year={2018}
}

@inproceedings{gu2017badnets,
  title={BadNets: Identifying Vulnerabilities in the Machine Learning Model Supply Chain},
  author={Gu, Tianyu and Dolan-Gavitt, Brendan and Garg, Siddharth},
  booktitle={Neural Information Processing Systems Workshop on Machine Learning Security},
  year={2017}
}

@article{tang2023cleanLabel,
  title={Did You Train on My Dataset? Towards Public Dataset Protection with CleanLabel Backdoor Watermarking},
  author={Tang, Ruixiang and Feng, Qizhang and Liu, Ninghao and Yang, Fan and Hu, Xia},
  journal={ACM SIGKDD Explorations Newsletter},
  volume={25},
  number={1},
  pages={43--53},
  year={2023}
}

@inproceedings{li2022untargeted,
  title={Untargeted Backdoor Watermark: Towards Harmless and Stealthy Dataset Copyright Protection},
  author={Li, Yiming and Bai, Yang and Jiang, Yong and Yang, Yong and Xia, Shu-Tao and Li, Bo},
  booktitle={Advances in Neural Information Processing Systems},
  volume={35},
  pages={13238--13250},
  year={2022}
}

@article{turner2019label,
  title={Label-consistent backdoor attacks},
  author={Turner, Alexander and Tsipras, Dimitris and Madry, Aleksander},
  journal={arXiv preprint arXiv:1912.02771},
  year={2019}
}

@inproceedings{sablayrolles2020radioactive,
  title={Radioactive data: tracing through training},
  author={Sablayrolles, Alexandre and Douze, Matthijs and Schmid, Cordelia and J{\'e}gou, Herv{\'e}},
  booktitle={International Conference on Machine Learning},
  pages={8326--8335},
  year={2020}
}

@inproceedings{guo2024domain,
  title={Domain Watermark: Effective and Harmless Dataset Copyright Protection is Closed at Hand},
  author={Guo, Junfeng and Li, Yiming and Wang, Lixu and Xia, Shu-Tao and Huang, Heng and Liu, Cong and Li, Bo},
  booktitle={Advances in Neural Information Processing Systems},
  volume={36},
  year={2024}
}

@inproceedings{wu2022watermarking,
  title={Watermarking Pre-trained Encoders in Contrastive Learning},
  author={Wu, Yutong and Qiu, Han and Zhang, Tianwei and Qiu, Meikang},
  booktitle={International Conference on Machine Learning},
  year={2022}
}

@inproceedings{cong2022sslguard,
  title={SSLGuard: A Watermarking Scheme for Self-supervised Learning Pre-trained Encoders},
  author={Cong, Tianshuo and He, Xinlei and Zhang, Yang},
  booktitle={ACM SIGSAC Conference on Computer and Communications Security},
  pages={579--593},
  year={2022}
}

@inproceedings{zhu2025evading,
  title={Evading Data Provenance in Deep Neural Networks},
  author={Zhu, Hongyu and Liang, Sichu and Wang, Wenwen and Zhang, Zhuomeng and Li, Fangqi and Wang, Shi-Lin},
  booktitle={IEEE/CVF International Conference on Computer Vision},
  pages={1249--1260},
  year={2025}
}

@article{shao2025databench,
  title={DATABench: Evaluating Dataset Auditing in Deep Learning from an Adversarial Perspective},
  author={Shao, Shuo and Li, Yiming and Zheng, Mengren and Hu, Zhiyang and Chen, Yukun and Li, Boheng and He, Yu and Guo, Junfeng and Tao, Dacheng and Qin, Zhan},
  journal={arXiv preprint arXiv:2507.05622},
  year={2025}
}

@article{wang2025sscl,
  title={SSCL-BW: Sample-Specific Clean-Label Backdoor Watermarking for Dataset Ownership Verification},
  author={Wang, Yingjia and Qiao, Ting and Liu, Xing and Li, Chongzuo and Wu, Sixing and Li, Jianbin},
  journal={arXiv preprint arXiv:2510.26420},
  year={2025}
}

@article{nguyen2021wanet,
  title={Wanet--imperceptible warping-based backdoor attack},
  author={Nguyen, Anh and Tran, Anh},
  journal={arXiv preprint arXiv:2102.10369},
  year={2021}
}

@article{chen2017targeted,
  title={Targeted backdoor attacks on deep learning systems using data poisoning},
  author={Chen, Xinyun and Liu, Chang and Li, Bo and Lu, Kimberly and Song, Dawn},
  journal={arXiv preprint arXiv:1712.05526},
  year={2017}
}

@article{bojanowski2017optimizing,
  title={Optimizing the latent space of generative networks},
  author={Bojanowski, Piotr and Joulin, Armand and Lopez-Paz, David and Szlam, Arthur},
  journal={arXiv preprint arXiv:1707.05776},
  year={2017}
}

@article{denton2015deep,
  title={Deep generative image models using a laplacian pyramid of adversarial networks},
  author={Denton, Emily L and Chintala, Soumith and Fergus, Rob and others},
  journal={Advances in neural information processing systems},
  volume={28},
  year={2015}
}

@inproceedings{ronneberger2015u,
  title={U-net: Convolutional networks for biomedical image segmentation},
  author={Ronneberger, Olaf and Fischer, Philipp and Brox, Thomas},
  booktitle={International Conference on Medical image computing and computer-assisted intervention},
  pages={234--241},
  year={2015},
  organization={Springer}
}

@inproceedings{muhammad2020eigen,
  title={Eigen-cam: Class activation map using principal components},
  author={Muhammad, Mohammed Bany and Yeasin, Mohammed},
  booktitle={2020 international joint conference on neural networks (IJCNN)},
  pages={1--7},
  year={2020},
  organization={IEEE}
}

@inproceedings{zhang2018unreasonable,
  title={The unreasonable effectiveness of deep features as a perceptual metric},
  author={Zhang, Richard and Isola, Phillip and Efros, Alexei A and Shechtman, Eli and Wang, Oliver},
  booktitle={Proceedings of the IEEE conference on computer vision and pattern recognition},
  pages={586--595},
  year={2018}
}

@inproceedings{irvin2019chexpert,
  title={Chexpert: A large chest radiograph dataset with uncertainty labels and expert comparison},
  author={Irvin, Jeremy and Rajpurkar, Pranav and Ko, Michael and Yu, Yifan and Ciurea-Ilcus, Silviana and Chute, Chris and Marklund, Henrik and Haghgoo, Behzad and Ball, Robyn and Shpanskaya, Katie and others},
  booktitle={Proceedings of the AAAI conference on artificial intelligence},
  volume={33},
  pages={590--597},
  year={2019}
}

@inproceedings{liu2017neural,
  title={Neural trojans},
  author={Liu, Yuntao and Xie, Yang and Srivastava, Ankur},
  booktitle={2017 IEEE international conference on computer design (ICCD)},
  pages={45--48},
  year={2017},
  organization={IEEE}
}

@inproceedings{liu2018fine,
  title={Fine-pruning: Defending against backdooring attacks on deep neural networks},
  author={Liu, Kang and Dolan-Gavitt, Brendan and Garg, Siddharth},
  booktitle={International symposium on research in attacks, intrusions, and defenses},
  pages={273--294},
  year={2018},
  organization={Springer}
}

@article{rajpurkar2023current,
  title={The current and future state of AI interpretation of medical images},
  author={Rajpurkar, Pranav and Lungren, Matthew P},
  journal={New England Journal of Medicine},
  volume={388},
  number={21},
  pages={1981--1990},
  year={2023},
  publisher={Mass Medical Soc}
}

@article{ma2024segment,
  title={Segment anything in medical images},
  author={Ma, Jun and He, Yuting and Li, Feifei and Han, Lin and You, Chenyu and Wang, Bo},
  journal={Nature Communications},
  volume={15},
  number={1},
  pages={654},
  year={2024},
  publisher={Nature Publishing Group UK London}
}

@article{diaz2024monai,
  title={Monai label: A framework for ai-assisted interactive labeling of 3d medical images},
  author={Diaz-Pinto, Andres and Alle, Sachidanand and Nath, Vishwesh and Tang, Yucheng and Ihsani, Alvin and Asad, Muhammad and P{\'e}rez-Garc{\'\i}a, Fernando and Mehta, Pritesh and Li, Wenqi and Flores, Mona and others},
  journal={Medical Image Analysis},
  volume={95},
  pages={103207},
  year={2024},
  publisher={Elsevier}
}

@article{ebrahimian2022fda,
  title={FDA-regulated AI algorithms: trends, strengths, and gaps of validation studies},
  author={Ebrahimian, Shadi and Kalra, Mannudeep K and Agarwal, Sheela and Bizzo, Bernardo C and Elkholy, Mona and Wald, Christoph and Allen, Bibb and Dreyer, Keith J},
  journal={Academic radiology},
  volume={29},
  number={4},
  pages={559--566},
  year={2022},
  publisher={Elsevier}
}

@article{johnson2023mimic,
  title={MIMIC-IV, a freely accessible electronic health record dataset},
  author={Johnson, Alistair EW and Bulgarelli, Lucas and Shen, Lu and Gayles, Alvin and Shammout, Ayad and Horng, Steven and Pollard, Tom J and Hao, Sicheng and Moody, Benjamin and Gow, Brian and others},
  journal={Scientific data},
  volume={10},
  number={1},
  pages={1},
  year={2023},
  publisher={Nature Publishing Group UK London}
}

@article{johnson2019mimic,
  title={MIMIC-CXR, a de-identified publicly available database of chest radiographs with free-text reports},
  author={Johnson, Alistair EW and Pollard, Tom J and Berkowitz, Seth J and Greenbaum, Nathaniel R and Lungren, Matthew P and Deng, Chih-ying and Mark, Roger G and Horng, Steven},
  journal={Scientific data},
  volume={6},
  number={1},
  pages={317},
  year={2019},
  publisher={Nature Publishing Group UK London}
}

@article{hua2023unambiguous,
  title={Unambiguous and high-fidelity backdoor watermarking for deep neural networks},
  author={Hua, Guang and Teoh, Andrew Beng Jin and Xiang, Yong and Jiang, Hao},
  journal={IEEE Transactions on Neural Networks and Learning Systems},
  volume={35},
  number={8},
  pages={11204--11217},
  year={2023},
  publisher={IEEE}
}

@inproceedings{feng2022fiba,
  title={Fiba: Frequency-injection based backdoor attack in medical image analysis},
  author={Feng, Yu and Ma, Benteng and Zhang, Jing and Zhao, Shanshan and Xia, Yong and Tao, Dacheng},
  booktitle={Proceedings of the IEEE/CVF Conference on Computer Vision and Pattern Recognition},
  pages={20876--20885},
  year={2022}
}

@inproceedings{jabbour2020deep,
  title={Deep learning applied to chest X-rays: exploiting and preventing shortcuts},
  author={Jabbour, Sarah and Fouhey, David and Kazerooni, Ella and Sjoding, Michael W and Wiens, Jenna},
  booktitle={Machine Learning for Healthcare Conference},
  pages={750--782},
  year={2020},
  organization={PMLR}
}

\end{document}